\documentclass[runningheads]{llncs}

\usepackage[T1]{fontenc}
\usepackage{graphicx}
\usepackage{booktabs}
\usepackage{array}
\usepackage{amsmath}
\usepackage{amssymb}
\usepackage{multirow}
\usepackage{float}

\title{AutoViewMem: Self-Configuring Orthogonal Views for Conversational Long-Term Memory}
\titlerunning{AutoViewMem}

\author{Zijie Cao \and Xijun Qu \and Zhicheng Gu \and
Xiaoshu Chen \and Duanyang Yuan \and Yanning Hou \and
Sihang Zhou\thanks{Corresponding author.} \and Jianxing Gong \and
Jian Huang \and Yang Mei\textsuperscript{*}}
\authorrunning{Z. Cao et al.}
\institute{National University of Defense Technology, Changsha, China \\
\email{\{caozijie,zhousihang12,yangmei\}@nudt.edu.cn}}

\begin{document}

\maketitle

\begin{abstract}
Long-term memory is essential for large language model (LLM) agents to maintain consistency and personalization over extended interactions. Existing memory systems typically rely on fixed granularities or static schemas, but these designs struggle when heterogeneous information---such as preferences, events, constraints, and temporal updates---is embedded in a single mixed representation. The resulting semantic interference makes top-$K$ retrieval sensitive to noise and often leaves relevant evidence poorly ranked.
We present AutoViewMem, a data-driven framework that organizes long-term conversational memory into self-configuring, low-overlap semantic views before indexing. AutoViewMem discovers candidate views from interaction traces, selects a compact complementary view set, and uses these views to guide write-time structured extraction of provenance-grounded memories. This representation-first design moves semantic disentanglement from retrieval time to write time, allowing standard top-$K$ similarity search to retrieve focused evidence without explicit routing or iterative retrieval. We further apply offline consolidation to improve memory compactness and consistency.
Experiments on the LoCoMo and PersonaMem benchmarks, under both Qwen3-8B and Qwen3-14B backbones, show that AutoViewMem improves long-horizon question answering and personalization over strong memory baselines while preserving a simple inference pipeline.

\end{abstract}

\section{Introduction}

 Large language model (LLM)-driven conversational agents are increasingly deployed in long-term companionship, personalized assistance, and enterprise support \cite{zhong2024memorybank,maharana2024evaluating,packer2023memgpt}. In these scenarios, interactions spanning weeks or months quickly exceed any finite context window. Even with long-context models, naively supplying the full history is rarely effective: topic drift, evolving user states, and a long-tail distribution of salient facts mean that large amounts of low-relevance context increase cost and noise while crucial details remain buried in distant tokens \cite{liu2024lost,maharana2024evaluating,wang2023longmem}.

 To address this, recent systems adopt external non-parametric memory under a write--retrieve--generate paradigm \cite{lewis2020retrieval,zhong2024memorybank,packer2023memgpt,wang2023longmem,chhikara2025mem0}. The central design problem is often framed as memory granularity: fine-grained units fragment evidence and hurt recall, while coarse-grained units inject irrelevant content. This has motivated multi-granularity methods such as MemGAS \cite{xu2025memgas} and segmentation-and-compression pipelines such as SeCom \cite{pan2025secom}.

However, granularity alone does not fully explain retrieval failures. Long conversations are intrinsically heterogeneous---preferences, plans, facts, events, and temporal updates often coexist in the same dialogue span. When such mixed-content memories are embedded in a single representation space, semantically unrelated information interferes during similarity search, creating blind spots and a persistent coverage--noise trade-off \cite{liu2024lost,maharana2024evaluating,pan2025secom,xu2025memgas,chhikara2025mem0,hu2026evermembench}. This points to a representation question that arises \emph{before} retrieval: how should memory be organized before indexing so that heterogeneous semantics remain accessible under diverse future queries?

 We propose \textbf{AutoViewMem} (Figure~\ref{fig:architecture}), a data-driven framework that organizes long-term memory into self-configuring, low-overlap semantic views. A view is orthogonal to granularity: granularity controls memory unit size, whereas a view defines a semantic projection over the same evidence. AutoViewMem discovers candidate views from interaction traces, selects a compact complementary set, and uses these views to guide write-time structured extraction with timestamps and provenance. The key design choice is to reduce semantic entanglement before items enter the vector index---once memories are written under complementary views, a simple global top-$K$ search can construct focused contexts without explicit routing or iterative retrieval. An offline graph-based consolidation step further improves compactness and consistency.

 We evaluate AutoViewMem on the LoCoMo \cite{maharana2024evaluating} and PersonaMem \cite{jiang2025personamem} benchmarks. Results show that AutoViewMem improves answer quality over representative long-term memory baselines in several settings, including Mem0 \cite{chhikara2025mem0}, MemGAS \cite{xu2025memgas}, MemoryBank \cite{zhong2024memorybank}, and A-mem \cite{xu2025amem}. The gains are supported by ablations and retrieval diagnostics showing that low-overlap multi-view organization improves the ranking and separation of relevant evidence under a fixed retrieval budget.

Our main contributions are:
\begin{itemize}
    \item \textbf{A representation-first perspective on conversational memory,} arguing that semantic interference in mixed-content representations is as central as granularity to long-horizon retrieval failures.
    \item \textbf{AutoViewMem,} a self-configuring framework that discovers complementary low-overlap views from streaming interactions and uses them to structure memory before indexing, enabling effective top-$K$ retrieval without complex query-time control.
    \item \textbf{Empirical evidence} across LoCoMo and PersonaMem, with ablations and retrieval diagnostics, that write-time view-based organization improves long-horizon QA and personalization over representative baselines.
\end{itemize}
\section{Related Work}

\paragraph{Retrieval and long-term memory.}
Retrieval grounds LLM outputs in external evidence: sparse retrievers such as BM25 remain strong lexical baselines \cite{robertson2009bm25}, while dense retrieval, contrastive retrievers, and RAG-style pipelines make semantic matching the standard interface for external memory \cite{karpukhin2020dpr,lewis2020retrieval,izacard2022contriever,luo2024bgelandmark}. Building on this, a growing body of work equips LLM agents with long-term memory through memory paging, external modules, time-aware updating, hierarchical storage, and scalable consolidation \cite{packer2023memgpt,wang2023longmem,zhong2024memorybank,chhikara2025mem0,kang2025memoryos,xu2025amem,fang2025lightmem,zhang2026emem,li2026timem,chen2025thinking,chen2025distilling,chen2026imgcot}. These systems mainly focus on how memory is stored, consolidated, or accessed at scale. Our work is complementary: rather than designing a new retriever or memory manager, we study how memory should be \emph{represented before indexing}, since retrieval quality in long-horizon dialogue depends on whether stored items preserve semantically focused evidence rather than entangled mixed-content spans \cite{liu2024lost,maharana2024evaluating,pan2025secom,xu2025memgas}. This is salient for agent memory, where the storage distribution is created online by the system itself rather than given as a fixed corpus.

\paragraph{Structured memory, schema induction, and benchmarks.}
Our work also relates to schema induction and structured memory construction. Prior work studies how reusable structures can be induced from unstructured text \cite{chambers2010schemas,li2023hierarchical}, and that slot structures can be induced directly from dialogue without manual design \cite{finch2024slot}; in conversational memory, entity--attribute records and preference frames are often more reusable than raw logs under topic drift \cite{zhong2024memorybank,pan2025secom}. Unlike approaches with fixed templates or manually specified fields, AutoViewMem induces lightweight semantic views online and uses them to guide write-time extraction while keeping retrieval simple. On the benchmark side, LoCoMo provides a controlled testbed for very long-term dyadic dialogue \cite{maharana2024evaluating}, and newer benchmarks extend toward multi-party, cross-topic, and temporally evolving settings \cite{hu2026evermembench,li2026locomoplus}, motivating evaluation of both final QA quality and the evidence-ranking behavior of the memory store.
\section{Methodology} \label{sec:method}

\begin{figure}[t]
    \centering
    \includegraphics[width=0.92\textwidth]{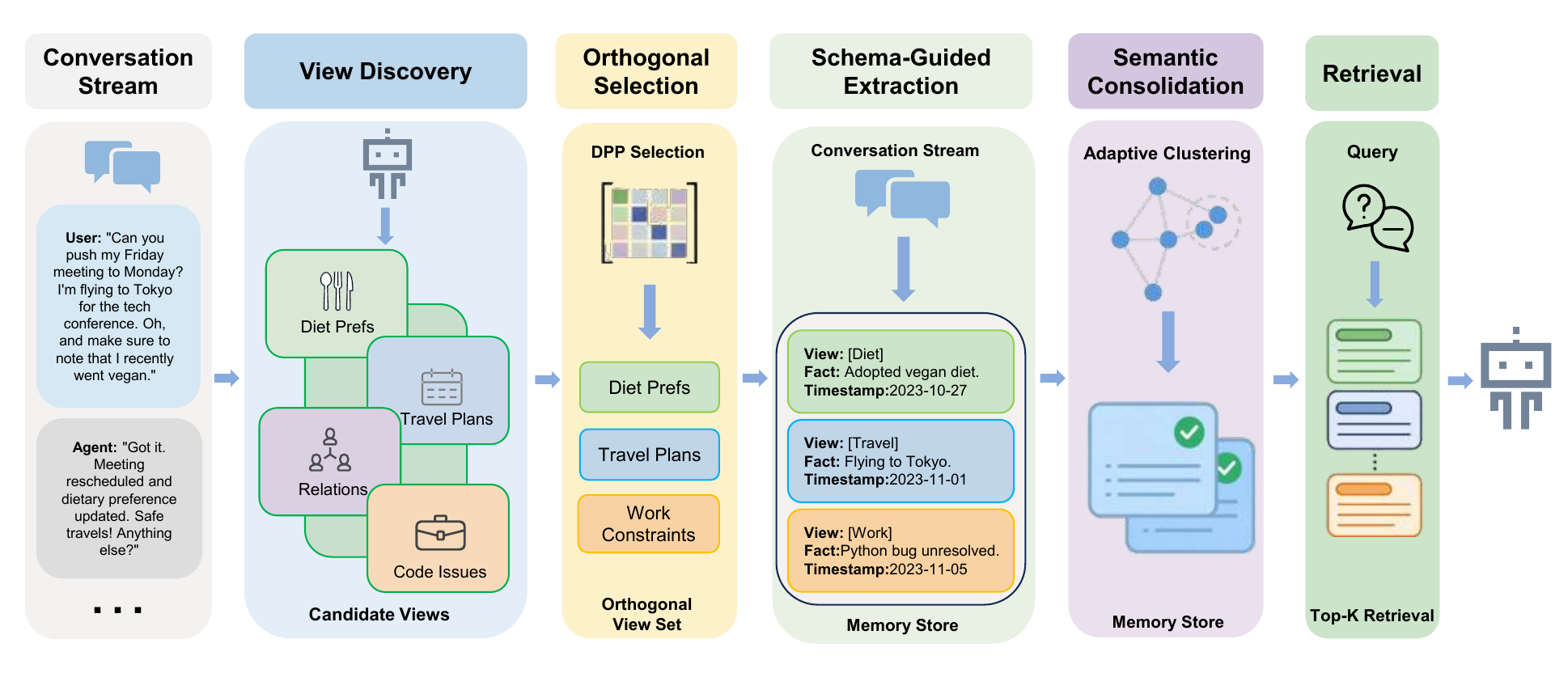}
    \caption{The AutoViewMem architecture. The system processes a conversation stream through online view discovery, selects an orthogonal view set via DPP, and performs schema-guided memory extraction and retrieval to provide context for the LLM agent.}
    \label{fig:architecture}
\end{figure}

We propose AutoViewMem (Figure~\ref{fig:architecture}), a two-stage memory architecture with (i) an online view induction-and-writing pipeline and (ii) offline consolidation for compactness and consistency. Online, the system incrementally induces lightweight reusable schemas---\emph{views}---from a conversation stream and uses them to guide structured memory writing. Offline, it deduplicates and consolidates extracted memories through graph clustering and LLM-based adjudication. The query-time interface remains conventional: the retriever searches a single index, while the main organization work happens before indexing.

\paragraph{Views and memory items.}
A view $v$ is a lightweight schema with a name, a set of slots, and an extraction template that instructs the LLM to populate those fields from dialogue \cite{pan2025secom,zhong2024memorybank}. We use \emph{orthogonal} pragmatically: views are intended to be complementary and low-overlap under semantic similarity, not strictly orthogonal in the linear-algebraic sense. For each user $u$ with conversation stream $\mathcal{S}_u$, we maintain an active view set $\mathcal{V}_u^*$ and a structured memory store $\mathcal{M}_u$ where each item carries its source span, timestamp, and view tag. Organizing memory under multiple focused views reduces the semantic entanglement that arises when heterogeneous content from the same dialogue span is stored as one undifferentiated representation.

\subsection{Online Adaptive Memory Learning} \label{subsec:online}

The online learner follows a Divergence--Convergence strategy with three phases: (i) \textbf{View Discovery}, which accumulates weak signals from the stream before committing to a stable schema basis; (ii) \textbf{View Convergence}, which aggregates noisy candidates into a compact active view set; and (iii) \textbf{Schema-guided Extraction}, which writes structured memories conditioned on the selected views.

\subsubsection{Phase 1: View Discovery} \label{subsubsec:phase1}

The conversation arrives as an ordered sequence of dialogue chunks. Instead of treating each chunk as an isolated memory write, the system first accumulates weak signals such as repeated entities, stable preferences, and recurring constraints. Periodically, every $N=50$ chunks, an LLM proposes $10$ candidate views $\tilde{\mathcal{V}}_u=\{v_1,\dots,v_{10}\}$ from the recent interaction trace. Each candidate view contains a name, slots, and an extraction template, making the candidate directly usable as a write-time extraction instruction.

\subsubsection{Phase 2: View Convergence} \label{subsubsec:phase2}

The candidate set $\tilde{\mathcal{V}}_u$ is noisy and redundant. We normalize exact-duplicate instructions and encode each remaining candidate with the same dense encoder used by the memory index. We select a compact, diverse active set $\mathcal{V}_u^*$ of $K=10$ views using a Determinantal Point Process (DPP) \cite{kulesza2012dpp,chen2018fastdpp} that discourages views inducing similar extraction behavior; the non-DPP ablation (Section~\ref{subsec:ablation}) uses uniform random selection. For each view $v_i$ with $\ell_2$-normalized embedding $z_i$, the $L$-ensemble kernel is $L_{ij} = q_i (z_i^\top z_j) q_j$ with uniform quality $q_i$, and we favor subsets with large $\det(L_S)$. Each selected seed is expanded with its top-30 nearest neighbors and summarized by the LLM into a canonical extraction instruction.

\paragraph{View normalization.}
Candidate views are deduplicated at the instruction-string level, and semantically similar candidates are rewritten into one concise extraction instruction. Every extracted item stores its view tag, timestamp, and provenance. Views are not forced to be mutually exclusive; overlapping projections are handled downstream by deduplicating on provenance identity (see Section~\ref{subsec:offline}).

\subsubsection{Phase 3: Adaptive Extraction} \label{subsubsec:phase3}

Given $\mathcal{V}_u^*$, for each dialogue chunk $x$, the LLM evaluates the chunk under each active view and returns structured facts only when evidence is relevant to that view---allowing the same evidence to be projected into multiple complementary views when appropriate. The resulting items carry structured fields, view tags, timestamps, and provenance pointers.

At query time, we perform standard top-$K$ dense retrieval over all memory items in a single unified index \cite{lewis2020retrieval,karpukhin2020dpr}. Retrieved items are merged under a token budget with lightweight deduplication before being passed to the LLM. When multiple retrieved items share the same underlying evidence, only the first occurrence contributes to retrieval metrics, preventing multi-view projections from artificially inflating coverage.

\subsection{Offline Memory Consolidation} \label{subsec:offline}

Online extraction prioritizes responsiveness and recall, while offline consolidation improves compactness and semantic consistency. The stage is deliberately separated from online writing: it runs asynchronously and does not change the query-time retrieval interface.

We first remove exact duplicates by hashing canonicalized text and/or structured representations, keeping one record per duplicate and merging metadata (view tags, occurrence counts, provenance). We then build an undirected similarity graph $G=(V,E)$ over items with an edge $(i,j)$ when $\cos(e_i,e_j)\ge 0.9$; connected components define candidate consolidation batches, and oversized components are recursively split by increasing $\tau$ (e.g., $0.90 \rightarrow 0.999$). Similarity only proposes candidates, never forces merging: records encoding conflicting claims, different time versions, or distinct concrete details remain separate unless the consolidation prompt can safely resolve them. Because embedding similarity alone is insufficient for safe merging, for each cluster an LLM rewrites the grouped items into one or more canonical records according to three cases---\textbf{Containment} (one item subsumes another), \textbf{Complementarity} (non-overlapping facts merge), and \textbf{Independence despite similarity} (similar items expressing different facts stay separate) \cite{zhong2024memorybank}. Canonical records retain merged provenance and replace redundant variants, reducing memory growth while preserving the evidence needed to audit or reverse a consolidation.
\section{Experiments}

We evaluate AutoViewMem on LoCoMo \cite{maharana2024evaluating} and PersonaMem \cite{jiang2025personamem}, testing whether multi-view write-time organization improves long-horizon QA, which components drive the gains, and whether the resulting store improves evidence ranking under a fixed retrieval budget.

\subsection{Experimental Setup}

\paragraph{Benchmarks and metrics.}
\textbf{LoCoMo} \cite{maharana2024evaluating} has 200--400 turn conversations with 1{,}540 questions spanning Multi-Hop, Temporal, Open-Domain, and Single-Hop reasoning; we report BLEU-1 (B1), F1, and LLM-Judge (J), category-wise and overall, using a unified Qwen3-8B judge. \textbf{PersonaMem} \cite{jiang2025personamem} is a personalization benchmark; we use its 32k-context tier (589 multiple-choice questions) and report accuracy under the official deterministic protocol, with per-capability breakdown.

\paragraph{Retrieval evaluation.}
AutoViewMem may produce multiple view-specific items grounded in the same evidence, so item-level retrieval can overestimate performance by rewarding redundant hits. We evaluate at the \emph{evidence level} using provenance pointers: each memory item records its source span IDs, and at evaluation we map retrieved items to provenance identity and match against ground-truth dialogue turns. For each query we retrieve top-$K_0=30$ items per speaker-specific store, merge, and deduplicate by evidence identity. We report the no-positive rate---the fraction of queries where no relevant evidence appears in the retrieved pool.

\paragraph{Backbones and baselines.}
We evaluate with Qwen3-8B and Qwen3-14B backbones, used for both memory processing and generation, against representative systems---A-mem \cite{xu2025amem}, Mem0 \cite{chhikara2025mem0}, MemGAS \cite{xu2025memgas}, and MemoryBank \cite{zhong2024memorybank}---and a Full-History oracle. All methods share the same embedding (e5-base-v2) and LLM backend, with temperature 0, max 8{,}192 generation tokens, and hyperparameters tuned on a held-out validation split.

\subsection{Main Results}

Tables~\ref{tab:main_results} and \ref{tab:personamem_results} report end-to-end results under both backbones.

On LoCoMo, AutoViewMem attains the best overall Judge under both backbones and the strongest overall F1 and BLEU-1, surpassing Full History on Judge under Qwen3-8B and trailing it only narrowly under Qwen3-14B. Lexical gains are most pronounced on Temporal and Single-Hop questions, indicating that view-organized memory yields more grounded answers.

On PersonaMem-32k, AutoViewMem attains the best overall accuracy under both backbones, outperforming the strongest baseline by over five points and Full History by nearly fourteen points under Qwen3-14B. Its advantage concentrates on capabilities requiring synthesis rather than mere recall---generalization, preference recommendation, suggestion, and recall of reasons---and widens with the larger backbone, suggesting more capable generators make better use of view-organized memory.

Since AutoViewMem uses standard top-$K$ retrieval, these gains stem primarily from how memories are written and consolidated rather than from a more complex query-time controller.

\begin{table}[H]
\centering
\caption{Main Results on LoCoMo (1{,}540 questions) under Qwen3-8B and Qwen3-14B generation, scored by a unified Qwen3-8B judge. Each category and the overall block report LLM-Judge (J), F1, and BLEU-1 (B1). Best results among external-memory methods are in \textbf{bold}; Full History is a full-context oracle (not bolded).}
\label{tab:main_results}
\scriptsize
\setlength{\tabcolsep}{3pt}
\renewcommand{\arraystretch}{1.0}
\resizebox{\textwidth}{!}{%
\begin{tabular}{l ccc ccc ccc ccc ccc}
\toprule
\multirow{2}{*}{\textbf{Method}} & \multicolumn{3}{c}{\textbf{Multi-Hop}} & \multicolumn{3}{c}{\textbf{Temporal}} & \multicolumn{3}{c}{\textbf{Open-Domain}} & \multicolumn{3}{c}{\textbf{Single-Hop}} & \multicolumn{3}{c}{\textbf{Overall}} \\
\cmidrule(lr){2-4} \cmidrule(lr){5-7} \cmidrule(lr){8-10} \cmidrule(lr){11-13} \cmidrule(lr){14-16}
& \textbf{J} & \textbf{F1} & \textbf{B1} & \textbf{J} & \textbf{F1} & \textbf{B1} & \textbf{J} & \textbf{F1} & \textbf{B1} & \textbf{J} & \textbf{F1} & \textbf{B1} & \textbf{J} & \textbf{F1} & \textbf{B1} \\
\midrule
\multicolumn{16}{c}{\textbf{Qwen3-8B}} \\
\midrule
Full History & 0.819 & 0.337 & 0.176 & 0.704 & 0.268 & 0.178 & 0.625 & 0.155 & 0.109 & 0.888 & 0.444 & 0.334 & 0.821 & 0.370 & 0.258 \\
A-mem & 0.713 & 0.276 & 0.150 & 0.536 & 0.361 & 0.226 & 0.646 & 0.122 & 0.086 & 0.801 & 0.435 & 0.307 & 0.720 & 0.371 & 0.248 \\
MemGAS & 0.752 & 0.157 & 0.081 & 0.741 & 0.107 & 0.036 & \textbf{0.740} & 0.104 & 0.042 & 0.815 & 0.229 & 0.125 & 0.783 & 0.183 & 0.093 \\
MemoryBank & 0.638 & 0.208 & 0.093 & 0.632 & 0.261 & 0.153 & 0.583 & 0.130 & 0.074 & 0.755 & 0.337 & 0.234 & 0.697 & 0.285 & 0.181 \\
Mem0 & 0.681 & 0.269 & 0.145 & 0.570 & 0.325 & 0.202 & 0.594 & 0.162 & 0.098 & 0.737 & 0.350 & 0.268 & 0.683 & 0.319 & 0.222 \\
AutoViewMem & \textbf{0.787} & \textbf{0.367} & \textbf{0.211} & \textbf{0.791} & \textbf{0.457} & \textbf{0.301} & 0.635 & \textbf{0.177} & \textbf{0.114} & \textbf{0.894} & \textbf{0.518} & \textbf{0.423} & \textbf{0.837} & \textbf{0.456} & \textbf{0.339} \\
\midrule
\addlinespace
\multicolumn{16}{c}{\textbf{Qwen3-14B}} \\
\midrule
Full History & 0.883 & 0.327 & 0.156 & 0.717 & 0.266 & 0.120 & 0.698 & 0.195 & 0.089 & 0.941 & 0.487 & 0.321 & 0.868 & 0.393 & 0.235 \\
A-mem & 0.694 & 0.291 & 0.151 & 0.704 & 0.297 & 0.182 & 0.660 & 0.148 & 0.093 & 0.801 & 0.470 & 0.383 & 0.751 & 0.379 & 0.278 \\
MemGAS & 0.766 & 0.122 & 0.057 & \textbf{0.844} & 0.074 & 0.025 & \textbf{0.833} & 0.088 & 0.035 & 0.847 & 0.187 & 0.096 & 0.831 & 0.145 & 0.070 \\
MemoryBank & 0.677 & 0.242 & 0.099 & 0.598 & 0.246 & 0.108 & 0.573 & 0.161 & 0.085 & 0.762 & 0.370 & 0.233 & 0.701 & 0.308 & 0.173 \\
Mem0 & 0.794 & 0.283 & 0.136 & 0.701 & 0.299 & 0.149 & 0.625 & 0.166 & 0.064 & 0.795 & 0.405 & 0.261 & 0.765 & 0.346 & 0.203 \\
AutoViewMem & \textbf{0.809} & \textbf{0.374} & \textbf{0.199} & 0.782 & \textbf{0.513} & \textbf{0.320} & 0.750 & \textbf{0.225} & \textbf{0.153} & \textbf{0.906} & \textbf{0.544} & \textbf{0.415} & \textbf{0.853} & \textbf{0.486} & \textbf{0.339} \\
\bottomrule
\end{tabular}
}
\end{table}

\begin{table}[ht]
\centering
\caption{Main Results on PersonaMem-32k (589 questions) under Qwen3-8B and Qwen3-14B generation. We report overall accuracy (\%) under the official deterministic multiple-choice protocol and per-capability accuracy: recall of user-shared facts (R-Fact), recall of mentioned content (R-Ment), preference tracking (Track), recall of reasons (R-Rsn), preference recommendation (P-Rec), generalization (Gen), and suggestion (Sugg). Best results among external-memory methods are in \textbf{bold}; Full History is a full-context oracle (not bolded).}
\label{tab:personamem_results}
\scriptsize
\setlength{\tabcolsep}{2pt}
\renewcommand{\arraystretch}{0.88}
\begin{tabular}{@{}lcccccccc@{}}
\toprule
\multirow{2}{*}{\textbf{Method}} & \multicolumn{7}{c}{\textbf{Per-capability accuracy}} & \multirow{2}{*}{\textbf{Acc}} \\
\cmidrule(lr){2-8}
& \textbf{R-Fact} & \textbf{R-Ment} & \textbf{Track} & \textbf{R-Rsn} & \textbf{P-Rec} & \textbf{Gen} & \textbf{Sugg} & \\
\midrule
\multicolumn{9}{c}{\textbf{Qwen3-8B}} \\
\midrule
Full History & 41.9 & 47.1 & 70.5 & 82.8 & 45.5 & 61.4 & 9.7 & 52.80 \\
A-mem & 72.1 & 70.6 & 60.4 & 77.8 & 50.9 & 54.4 & 18.3 & 58.06 \\
MemGAS & 65.9 & 58.8 & 62.6 & 68.7 & 43.6 & 29.8 & 8.6 & 50.76 \\
MemoryBank & \textbf{76.7} & \textbf{76.5} & 61.9 & 73.7 & 56.4 & 59.6 & 19.4 & 60.10 \\
Mem0 & 69.8 & 70.6 & 58.3 & 70.7 & 58.2 & 57.9 & 15.1 & 56.37 \\
AutoViewMem & 67.4 & 64.7 & \textbf{63.3} & \textbf{79.8} & \textbf{65.5} & \textbf{77.2} & \textbf{24.7} & \textbf{62.48} \\
\midrule
\addlinespace
\multicolumn{9}{c}{\textbf{Qwen3-14B}} \\
\midrule
Full History & 59.7 & 64.7 & 69.8 & 80.8 & 50.9 & 38.6 & 10.8 & 55.18 \\
A-mem & 66.7 & \textbf{76.5} & 67.6 & 80.8 & 65.5 & 73.7 & 23.7 & 63.33 \\
MemGAS & 56.6 & 52.9 & 64.8 & 73.7 & 50.9 & 42.1 & 11.8 & 52.29 \\
MemoryBank & \textbf{69.0} & 64.7 & \textbf{70.5} & 80.8 & 63.6 & 68.4 & 18.3 & 62.65 \\
Mem0 & 62.0 & 70.6 & 64.0 & \textbf{84.8} & 69.1 & 73.7 & 22.6 & 62.14 \\
AutoViewMem & 67.4 & 70.6 & 66.9 & 83.8 & \textbf{74.5} & \textbf{84.2} & \textbf{46.2} & \textbf{69.10} \\
\bottomrule
\end{tabular}
\end{table}

\subsection{Efficiency and Ablation Analysis}
\label{subsec:ablation}

\begin{figure}[ht]
    \centering
    \includegraphics[width=0.48\textwidth]{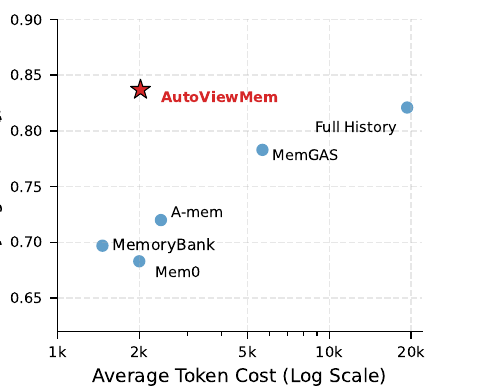}
    \hfill
    \includegraphics[width=0.48\textwidth]{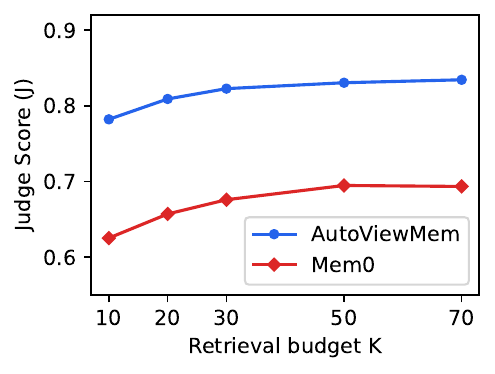}
    \caption{Left: performance--cost trade-off on LoCoMo (x-axis: average token cost, log scale; y-axis: Judge score). Right: effect of the top-$K$ retrieval budget on Judge score, comparing AutoViewMem with Mem0.}
    \label{fig:performance_vs_cost}
\end{figure}

\paragraph{Performance--cost trade-off.}
Figure~\ref{fig:performance_vs_cost} (left) compares answer quality and token cost. AutoViewMem improves quality over lower-cost baselines (Mem0, MemoryBank) at a comparable retrieval budget, indicating that better write-time organization increases context usefulness without expensive query-time control.

\paragraph{Ablation settings.}
We conduct ablations on the Qwen3-8B backbone: (1) \textit{Single-view Structured}, which removes multi-view organization while retaining structured extraction; (2) \textit{w/o DPP Selection}, which replaces DPP-based selection with random view selection; and (3) \textit{w/o Graph Consolidation}, which removes offline consolidation. These variants isolate whether the improvement comes from structure alone, diversity-aware view selection, or post-hoc memory cleanup.

\begin{table}[ht]
\centering
\caption{Ablation on the Qwen3-8B backbone. We report overall LLM-Judge (J), F1, BLEU-1 (B1), and the no-positive rate (no-pos\%, lower is better: the fraction of queries with no relevant evidence in the retrieved pool). All variants are evaluated under the same fixed $K_0=30$ retrieval budget.}
\label{tab:ablation}
\scriptsize
\setlength{\tabcolsep}{6pt}
\renewcommand{\arraystretch}{1.1}
\begin{tabular}{lcccc}
\toprule
\textbf{Variant} & \textbf{J} & \textbf{F1} & \textbf{B1} & \textbf{no-pos\%}$\downarrow$ \\
\midrule
AutoViewMem (Full) & 0.837 & 0.456 & 0.339 & 5.14 \\
\quad Single-view Structured & 0.774 & 0.422 & 0.317 & 5.53 \\
\quad w/o DPP Selection (random) & 0.777 & 0.428 & 0.322 & 7.03 \\
\quad w/o Graph Consolidation & 0.801 & 0.449 & 0.328 & 9.83 \\
\bottomrule
\end{tabular}
\end{table}

\paragraph{Ablation results.}
Table~\ref{tab:ablation} reports variants under a fixed $K_0=30$ budget. The full model attains the best Judge (0.837) and lowest no-positive rate (5.14\%), meaning it finds gold evidence for more queries. Removing multi-view organization gives the weakest Judge and BLEU-1, confirming structured extraction alone is insufficient. Removing consolidation drops J to 0.801 with the worst no-positive rate (9.83\%), showing redundancy control matters at a tight budget. The random-selection variant (\textit{w/o DPP}) is competitive on lexical metrics but has a higher no-positive rate (7.03\%) than the full model, consistent with DPP acting as a diversity prior rather than the sole source of gains.

\subsection{Retrieval Analysis}
\label{sec:retrieval_analysis}

\paragraph{Sensitivity to retrieval budget.}
Varying the top-$K$ context budget (Figure~\ref{fig:performance_vs_cost}, right), both AutoViewMem and Mem0 improve with $K$ and then saturate; AutoViewMem remains consistently better, especially at smaller budgets---focused views make early retrieved context less noisy.

\subsection{Case Study: User-Adaptive Semantic Projections of the Same Evidence}
\label{subsec:case_study}

Table~\ref{tab:case_study} illustrates how AutoViewMem differs from a fixed general-purpose memory extractor. The current utterance is intentionally ambiguous: it never explicitly mentions football or the user's previous difficulty finding partners. A fixed extraction prompt therefore preserves mainly the surface-level facts. In contrast, AutoViewMem interprets the same evidence through user-adaptive views induced from the interaction history, exposing multiple semantically distinct aspects of the utterance. Importantly, history-dependent interpretations are expressed conservatively when they are not directly confirmed by the current dialogue.

\paragraph{Source dialogue.}
``\emph{That grassy area is close to the office, and the lights are still on at ten. Li and the others finally agreed to come along. It doesn't matter if we're not that good; it would be nice just to chat afterward. At last, I won't have to do it alone every time.}''

\paragraph{Relevant interaction history.}
The user frequently plays football after work. Since moving, they have repeatedly mentioned difficulty finding people to play with and that practicing alone is less enjoyable.

\begin{table}[ht]
\centering
\caption{Case study on an ambiguous utterance. A fixed general-purpose prompt records only the surface-level content, whereas AutoViewMem's user-adaptive views project the same evidence onto three semantically distinct aspects (\textemdash{} = nothing exposed beyond surface extraction). The first projection is stated tentatively because the football reading is supported by the interaction history but is not confirmed in the current utterance.}
\label{tab:case_study}
\scriptsize
\setlength{\tabcolsep}{3pt}
\renewcommand{\arraystretch}{1.1}
\begin{tabular}{@{}p{0.18\textwidth} >{\raggedright\arraybackslash}p{0.19\textwidth} >{\raggedright\arraybackslash}p{0.29\textwidth} >{\raggedright\arraybackslash}p{0.28\textwidth}@{}}
\toprule
\textbf{Extraction} & \textbf{View} & \textbf{Extracted memory} & \textbf{Information exposed beyond surface extraction} \\
\midrule
\textbf{Fixed prompt} & General memory & A grassy area near the office remains lit at 10~p.m.\ Li and others have agreed to join the user. The user does not mind differences in skill level and would like to chat afterward. & \textemdash \\
\midrule
\multirow{3}{*}{\textbf{AutoViewMem}} & \textbf{Football activity \& venue} & The user may be considering the illuminated grassy area near the office as a place to play football with Li and others after work. & \textbf{Latent activity goal:} links venue, time, and participants to the user's recurring football activity, although football is not explicitly mentioned in the current utterance. \\
\addlinespace
 & \textbf{Social motivation} & For this activity, the user values participating together and socializing afterward more than differences in playing ability. & \textbf{Participation priority:} interprets tolerance of lower skill levels as a preference for companionship over competitive performance. \\
\addlinespace
 & \textbf{Companionship \& affect} & The user expresses relief that Li and others have agreed to join and looks forward to no longer practicing alone. & \textbf{Emotional significance:} connects \emph{``finally''} and \emph{``at last''} to the user's previously expressed frustration with solo practice. \\
\bottomrule
\end{tabular}
\end{table}

\paragraph{Interpretation.}
The fixed prompt mainly records \emph{what was explicitly said}. AutoViewMem additionally separates \emph{what the situation is about}, \emph{what the user values in it}, and \emph{why it matters emotionally}. These memories are not produced by simply storing more text: they arise from applying different user-adaptive semantic projections to the same dialogue evidence. The first projection is marked as tentative because the football interpretation is supported by interaction history but is not explicitly confirmed in the current utterance.
\section{Conclusion}

We presented AutoViewMem, a self-configuring multi-view framework for long-term conversational memory. Beyond granularity, our work highlights memory organization as a key factor in long-horizon retrieval: storing heterogeneous content in a single representation space introduces semantic interference that destabilizes top-$K$ search. By organizing interaction streams into complementary low-overlap views at write time, AutoViewMem supports provenance-grounded extraction and effective retrieval with a simple pipeline, with offline consolidation further improving compactness.

Results on LoCoMo and PersonaMem under both Qwen3-8B and Qwen3-14B backbones show that this representation-first design improves over strong external-memory baselines. Analyses confirm low-overlap multi-view organization as the main source of improvement, enabling more stable retrieval without complex query-time control.

AutoViewMem has limitations: it depends on the underlying LLM for view discovery, extraction, and consolidation; view convergence runs periodically from buffered traces rather than fully online, which may slow adaptation under rapid distribution shift; offline consolidation can over-merge when subtle temporal changes should keep items separate, though provenance retention reduces this risk; and evaluation is limited to LoCoMo and PersonaMem, requiring broader validation on multi-party, multilingual, or safety-critical settings. We will release our code, prompts, and evaluation scripts upon acceptance.

\bibliographystyle{splncs04}
\bibliography{references}

\begin{credits}
\subsubsection{\discintname} The authors have no competing interests to declare
that are relevant to the content of this article.
\end{credits}

\end{document}